\documentclass[letterpaper, 10 pt, conference]{ieeeconf}  % Comment this line out if you need a4paper

\IEEEoverridecommandlockouts 
\usepackage{epsfig} % for postscript graphics files
\usepackage{mathptmx} % assumes new font selection scheme installed
\usepackage{times} % assumes new font selection scheme installed
\usepackage{amsmath} % assumes amsmath package installed
\usepackage{amssymb}  % assumes amsmath package installed
\usepackage{graphicx}
\usepackage{subcaption}
\usepackage{siunitx}
\usepackage[font=small,labelfont=bf]{caption}

\usepackage[nolist,nohyperlinks]{acronym}
\usepackage{mwe} % for example-image
\usepackage{cleveref}
\usepackage{float}

\floatstyle{plaintop}
\restylefloat{table}
\title{\LARGE \bf
Surface-based Manipulation Using Tunable Compliant Porous-Elastic Soft Sensing  
}

\author{Gayatri Indukumar$^{1,2}$, Muhammad Awais$^{1,3}$, Diana Cafiso$^{1}$ , Matteo Lo Preti$^{4}$, Lucia Beccai$^{1}$% <-this % stops a space
\thanks{$^{1}$Gayatri Indukumar, Muhammad Awais, Diana Cafiso, and Lucia Beccai are with Soft BioRobotics Perception Group at Istituto Italiano di Tecnologia, 16163, Genova, Italy}%
\thanks{$^{2}$Gayatri Indukumar is also with the University of Genoa, 16126, Genova, Italy}
\thanks{$^{3}$Muhammad Awais is also with the Biorobotics Institute, Scuola Superiore Sant’Anna, Pontedera, 56025, Italy}%
\thanks{$^{4}$Matteo Lo Preti is with the Soft Robotics Lab, National University of Singapore, 117575, Singapore}%
\thanks{Corresponding Author: Lucia Beccai(email: Lucia.Beccai@iit.it)}%
}

\begin{document}

\maketitle
\thispagestyle{empty}
\pagestyle{empty}

%%%%%%%%%%%%%%%%%%%%%%%%%%%%%%%%%%%%%%%%%%%%%%%%%%%%%%%%%%%%%%%%%%%%%%%%%%%%%%%%
\begin{abstract}

There is a growing need for soft robotic platforms that perform gentle, precise handling of a wide variety of objects. Existing surface-based manipulation systems, however, lack the compliance and tactile feedback needed for delicate handling. This work introduces the \ac{COPESS} integrated with inductive sensors for adaptive object manipulation and localised sensing. The design features a tunable lattice layer that simultaneously modulates mechanical compliance and sensing performance. By adjusting lattice geometry, both stiffness and sensor response can be tailored to handle objects with varying mechanical properties. Experiments demonstrate that by easily adjusting one parameter, the lattice density, from \SI{7}{\percent} to \SI{20}{\percent}, it is possible to significantly alter the sensitivity and operational force range (about \num{-23}\texttimes~and \num{+9}\texttimes~respectively). This approach establishes a blueprint for creating adaptive, sensorized surfaces where mechanical and sensory properties are co-optimized, enabling passive, yet programmable, delicate manipulation.

\end{abstract}

%%%%%%%%%%%%%%%%%%%%%%%%%%%%%%%%%%%%%%%%%%%%%%%%%%%%%%%%%%%%%%%%%%%%%%%%%%%%%%%%

\section{INTRODUCTION}
%%**-----------------------------------------------------------------------
%   •	General introduction to MOZART application and state of art
%   •	Motivate need for Sensorization and brief state of art in possible sensor options
%   •	Motivate inductive sensor choice
%   •	Introduce the concept of porous-elastic layer
%   •	Tie it all up together and explain the novelty
%   •	Contributions
%%**-----------------------------------------------------------------------
Soft handling systems offer a promising tool for expanding the applications of soft robotics to various areas, such as the food and packaging industry, through their ability to transport and manipulate fragile and heterogeneous objects \cite{wang2025food}. There are several examples of flexible surfaces, where the surface is deformed by actuation mechanisms. Different approaches are based on bellow-like modules, electrostatically driven actuators \cite{wang2017dielectric}, bioinspired dynamic deformation \cite{deng2016novel}, and pulley-and-belt-based linear actuators \cite{ingle2025soft}. Despite advances in actuation and manipulation to move heterogeneous objects, these systems often lack adequate sensing. Among the few sensorized examples, ArrayBot \cite{xue2024arraybot} deploys commercial force resistive sensors, and \cite{wang2025surface} uses vision to track the geometric center position and orientation of objects.
Nevertheless, opportunities remain to advance embedded sensorization in these systems. Tactile sensors present an excellent fit as they offer information regarding contact location, pressure distribution, object shape, and surface texture. Tactile sensors have seen a lot of advancements in recent years, exploiting various transduction approaches \cite{awais2024multiplexed}, \cite{yun2014polymer}. In this work, inductive soft sensors are selected because they balance sensitivity \cite{chen2025highly}, spatial resolution, and cost-effectiveness \cite{8234098}, and are relatively insensitive to dust and moisture, making them well-suited for large-scale modular robotic systems in fields such as food manipulation and packaging \cite{wang2018robust}. 

   \begin{figure*}[tb]
      \centering
  %--- First subfigure ---
  \begin{subfigure}[t]{0.45\textwidth}
    \includegraphics[width=\linewidth]{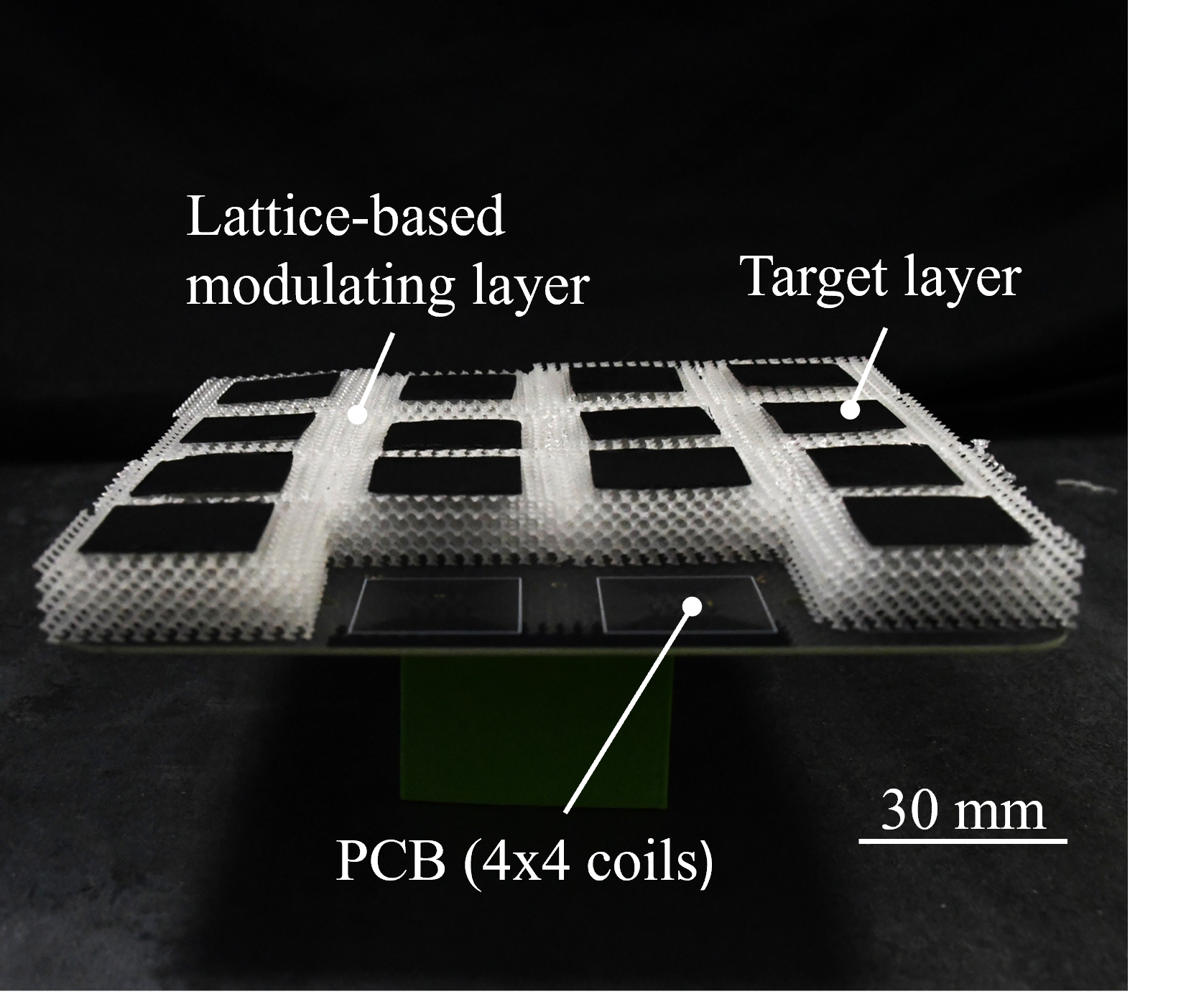}
    \caption{}
    \label{sfig:CIsys}
  \end{subfigure}
    \begin{subfigure}[t]{0.45\textwidth}
    \includegraphics[width=\linewidth]{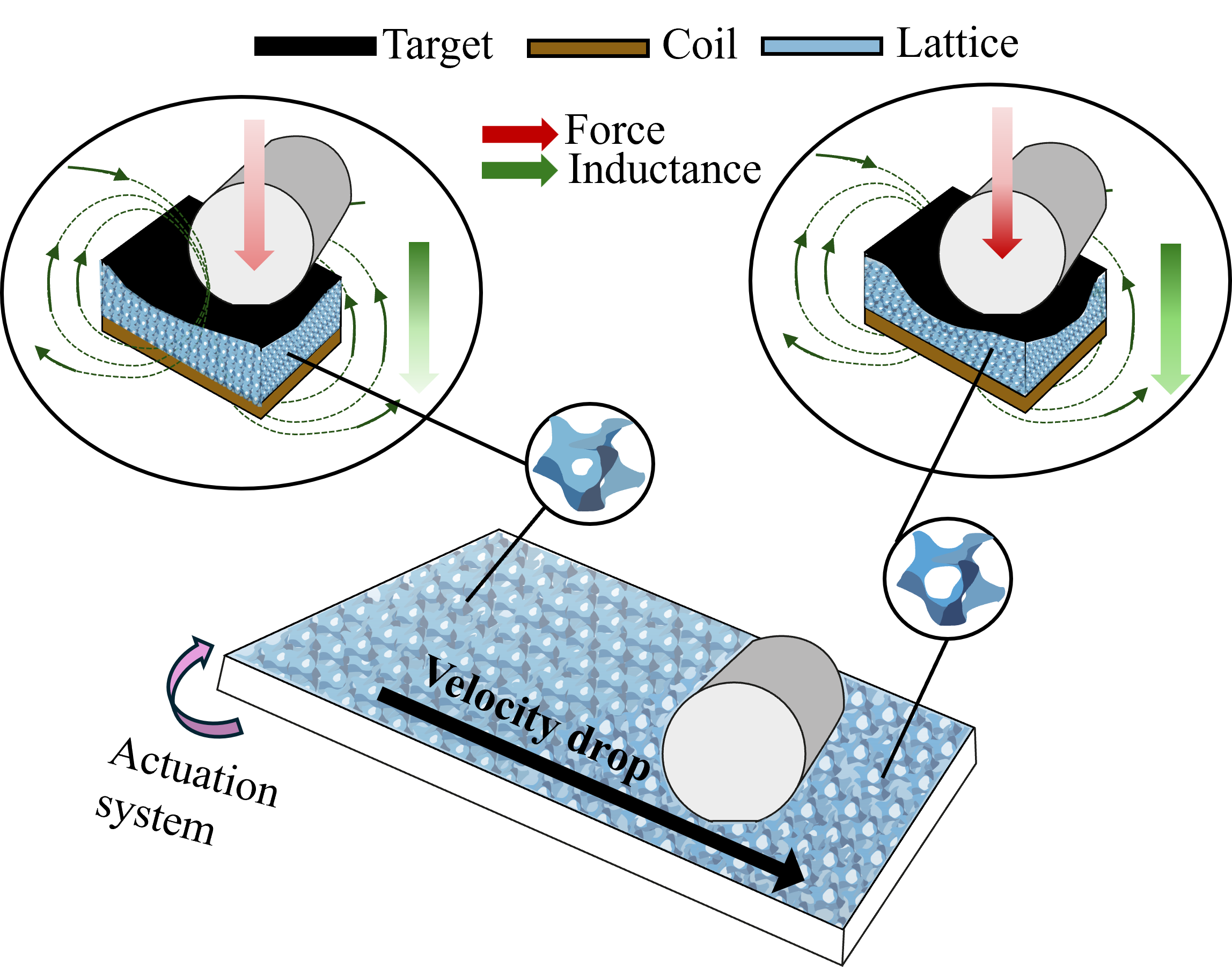}
    \caption{}
    \label{sfig:CIsch}
  \end{subfigure}
     \caption{Overview of the proposed COPESS-based surface manipulation system. (a) Photograph of the COPESS-based surface manipulation setup, consisting of a sensing unit with a 4×4 coil array, a lattice-based modulating layer, and a target layer. (b) Schematic illustration of the COPESS-based manipulation mechanism. The sensorized tile is covered with a 3D-printed lattice structure, which acts as a modulating layer. As the object transitions from a high-stiffness lattice to a low-stiffness lattice region, its velocity decreases while the sensor response increases.}
      \label{figure:CI}
   \end{figure*}
Among soft inductive tactile sensors, a variety of modulating layers are used. The most common choice is to use an elastomer like \ac{PDMS} or EcoFlex \cite{jones2019soft}, eventually combined with conductive fillers \cite{hamaguchi2020soft}, \cite{ozioko2019inductance}. Using a porous material represents another strategy to decrease the stiffness of the soft modulating layer. Porosity can be introduced via chemical routes \cite{cafiso2024dlp}, or by using commercial foams. In a previous work, we used a commercial polyurethane foam as the modulating layer \cite{dacre2025neural}. Although commercial foams are readily available and well-suited for application-driven studies, their mechanical properties cannot be tuned on demand. A practical compromise between tunability and rapid iteration is offered by 3D-printed lattices, which support fast prototyping across multiple designs.

%While lattices have been successfully used in resistive and capacitive tactile sensors \cite{cafiso2024dlp}, their application in inductive sensors has not yet been investigated. 

To address these challenges, we propose an approach that combines the environmental robustness of inductive sensing with the rapid design exploration offered by 3D-printed lattices. By integrating these two technologies, we create the \ac{COPESS}, a robust tactile surface capable of handling a wide variety of delicate and deformable objects.
%\ac{COPESS} is composed of flexible coils, a 3D-printed modulating layer, and a conductive target. 
%“porous-elasticity” in this context refers to the creation of deformable surfaces through 3D-printed lattices, designed to achieve different mechanical properties by easily changing the relative density. This enables fast tailoring of the sensor's mechanical properties, also across the platform. 
Building the layer with a gradient stiffness provides a smooth surface that delicately embeds objects, enabling precise tuning of the force-detection range and the potential to optimize manipulation along defined patterns.
The main contributions of this work are summarized as:
\begin{itemize}
    \item The investigation of 3D-printed gyroid lattices as a tunable modulating element for inductive tactile sensors.
    \item A comprehensive characterization demonstrating the co-modulation of mechanical stiffness and sensing range.
    \item A proof-of-concept demonstration that our approach can be used, in the future, for passive object guidance using programmed stiffness gradients.
\end{itemize}

\section{DESIGN}
%%**-----------------------------------------------------------------------
% Delve briefly into the theory of lattices
% Motivate the choice of printing method and material
% Motivate the choice of the lattice
    % Compare the lattice behavior with the standard plot and hypothesize the reason for the differences 
    % Compare against other choices
% Explain the fabrication process
    % How do we determine height

%%**----------------------------------------------------------------------
\Cref{figure:CI} shows the proposed scheme and a breakdown of the \acs{COPESS}. It consists of three layers: 
\begin{itemize}

\item Copper and ferrite are used as target layers in the inductive sensor, with copper placed close to the coil to generate strong eddy currents and ferrite positioned behind it to guide the magnetic flux and reduce leakage \cite{dong2025dual}. This configuration enhances inductance variation, improves sensitivity, and minimizes external interference.
\item The modulating layer consists of compliant lattices, which deform under the applied pressure, changing the distance between the coils and the target layer.
\item An array of 4x4 coils, made of four layers and with a dimension of \SI{25}{\milli \meter}x\SI{25}{\milli \meter}. The sensing tile consists of 16 coils, which are connected to 4 LDC1614 inductance-to-digital converter (Texas Instruments), which records the data with I2C protocol. The coils are excited by an \ac{AC} and generate a dynamic magnetic field. An additional layer of ferrite at the bottom is used to provide \ac{EM} shielding \cite{maass2017reduction}.  
 \end{itemize}

%The first step is to define the morphology of the modulating layer. 

% Gyroidal lattices can be defined using the following parametric equation:

% % \begin{equation}
% %      \sin x\cos y+\sin y\cos z+\sin z\cos x=0

\begin{figure}[tb]
  \centering
  %--- First subfigure ---
  \begin{subfigure}[t]{0.65\linewidth}
    \includegraphics[width=\linewidth]{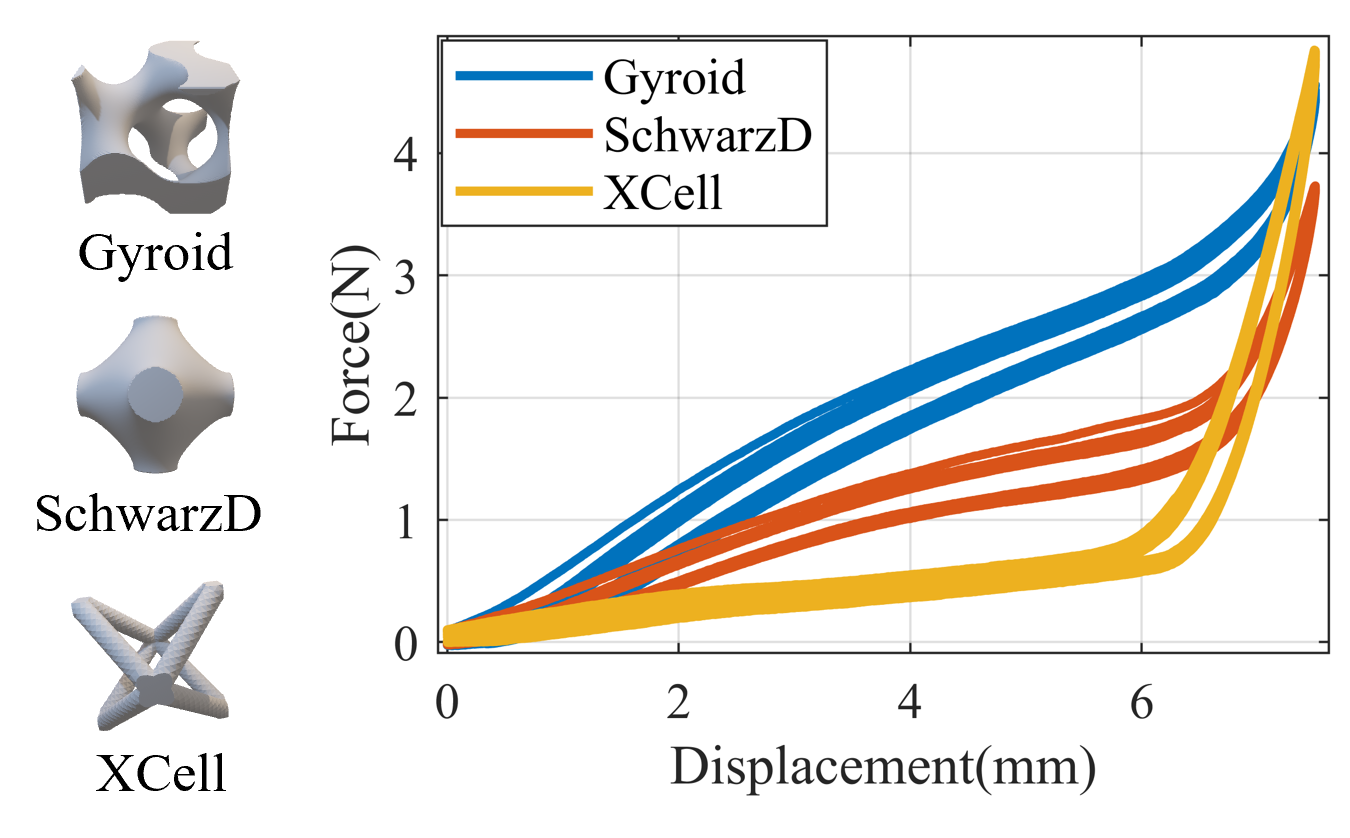}
    \caption{}
    \label{sfig:latcomp}
  \end{subfigure}

  % \vspace{0.5em}

  \begin{subfigure}[t]{0.5\linewidth}
    \includegraphics[width=\linewidth]{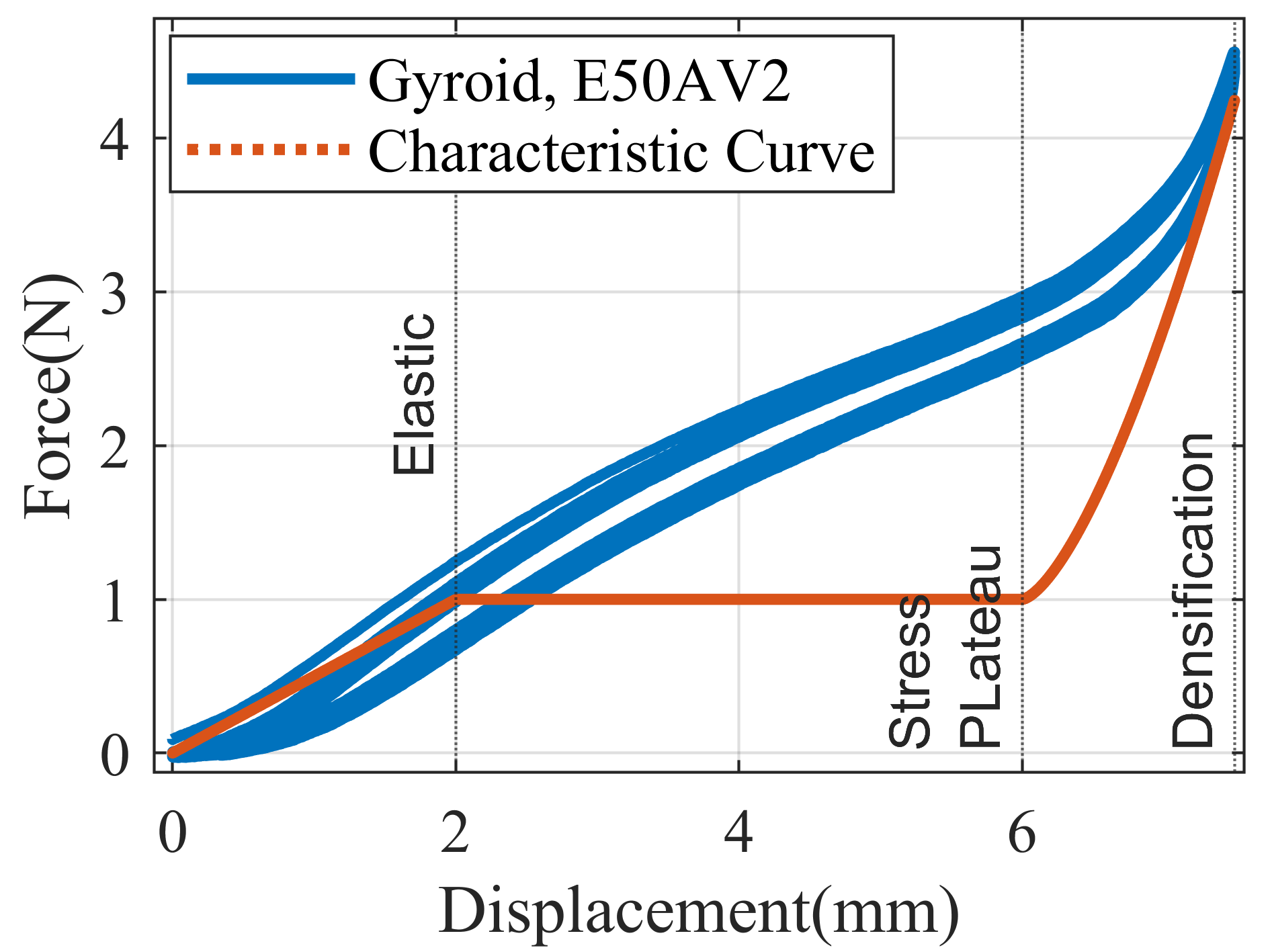}
    \caption{}
    \label{sfig:typcomp}
  \end{subfigure}

  % \vspace{0.5em}

  \begin{subfigure}[t]{0.5\linewidth}
    \includegraphics[width=\linewidth]{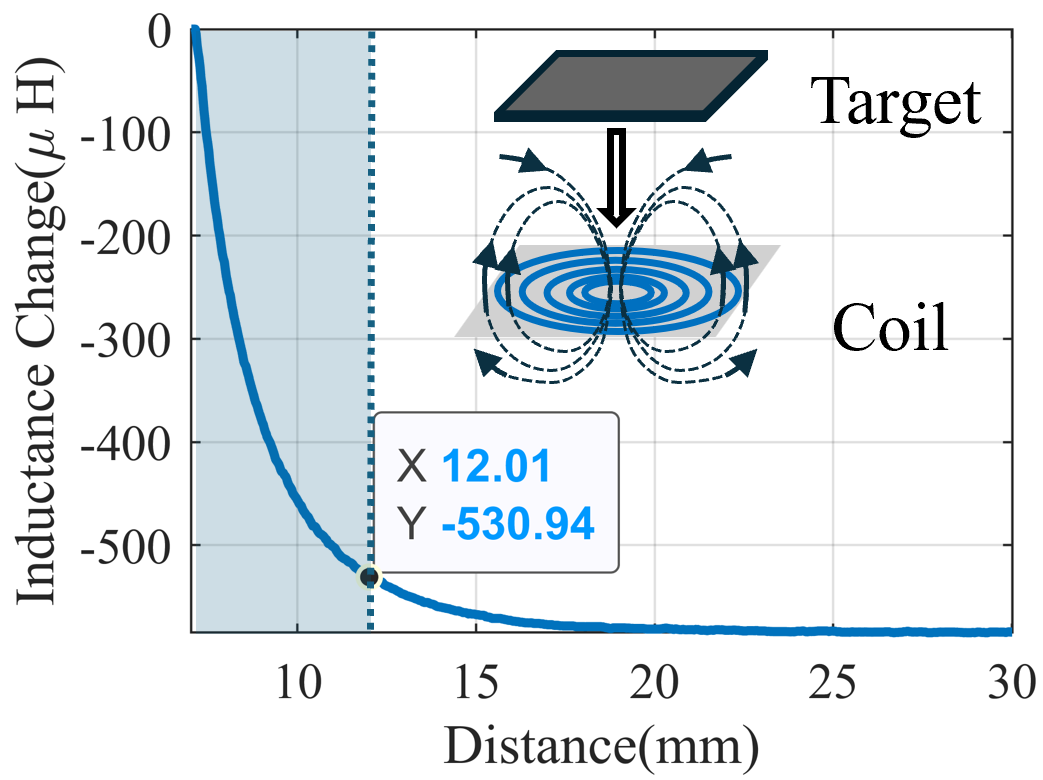}
    \caption{}
    \label{sfig:thickmot}
  \end{subfigure}

  \caption{Mechanical and electrical characterization informing the sensor's design. 
  (a) Force vs. displacement curves of three different lattice structures, demonstrating that the Gyroid (blue) offers the best combination of initial stiffness and strain range before densification. 
  (b) Comparing a gyroid lattice structure printed using Elastic 50AV2 resin against the characteristic curve of a typical hyperelastic material. 
  (c) Inductance change as a function of target distance, which heuristically determined the chosen thickness for the modulating lattice layer.}
  \label{figure:thickmotive}
\end{figure}

Compression tests (\SI{0.5}{\milli \meter /\second}, max strain = 50\%) are performed to choose the most suitable lattice structure. In particular, SchwarzD, Gyroid, and XCell strut-based lattices are compared. For a given set of characteristics (Cell size = \SI{3}{\milli \meter}, \SI{14.5}{\milli \meter}$\phi$x\SI{12}{\milli \meter}), the gyroid exhibits the highest stiffness and the largest strain range before densification (\Cref{sfig:latcomp}), suggesting a wider working range. \Cref{sfig:typcomp} shows the mechanical characterization of the gyroid compared to the typical curve seen for the lattices \cite{HaneyPoly2023TPMS}. Notably, it is observed that the stress–strain curve no longer exhibits the typical stress plateau. This extended region can be injectively mapped against a sensor response. Once the lattice approaches densification, the response is unreliable. Therefore, the sensors are designed to operate in the region preceding densification.

% In food processing, handling fragile items requires a delicate yet precise approach to prevent damage while ensuring efficiency. Inductive sensing-based soft surfaces offer a robust solution for detecting object position, pressure, and are able to provide high precision localized sensing. Unlike capacitive or resistive sensors, inductive sensing remains unaffected by moisture and contaminations, making it ideal for hygienic food handling [?]. Moreover, inductive transduction has the intrinsic advantage of allowing a sort of telesensing. 

The thickness of the modulating layer is a critical design parameter. Any given coil has an effective region where the \acs{EM} field is strong enough to interact with the target. To ensure a predictable and near-linear relationship between object displacement and sensor output, we design the sensor to operate within the linear region of the coil's inductance-distance curve. As characterized in \Cref{sfig:thickmot}, this linear region extends to approximately 12 mm. Therefore, we set the lattice thickness to \SI{12}{\milli \meter} to maximize the operational range while maintaining high linearity and sensitivity. The planar area is set to the size of the coils. Therefore, the lattices are designed to be \qtyproduct[product-units = single]{37.5 x 37.5 x 12}{\milli\metre}. 
\subsection{Lattice Fabrication and Post-Processing}
Vat Photopolymerization techniques, such as \ac{SLA}, are well-suited for printing highly intricate lattices due to their high-resolution light-based curing. In this work, \acs{SLA} (Formlabs 4) is used to print Elastic 50A V2 resin, selected for its good initial compliance (Shore hardness = 50A) and fast recovery from deformation.  The printing parameters are as follows: perimeter, model, and support fill regions receive an energy dose of \SI{38.40}{\milli \joule / \centi \meter \squared}, whereas the light intensity is set as \SI{11.5}{\milli \watt \centi \meter \squared}, layer thickness is set at \SI{0.1}{\milli \meter}. 
%The samples are printed flat on the build plate. 
Once printed, the samples are washed in isopropanol for \SI{20}{\minute} and post-cured via UV irradiation for \SI{45}{\minute} at \SI{70}{\celsius} in a Form Cure unit (Formlabs). 
Various combinations of the lattice parameters (cell size, relative density, thickness) are printed and characterized. However, it is observed that there are strict limits on the cell size (\SIrange {2.2}{4}{\milli \meter}) and relative density (\SIrange{7}{30}{\percent}) based on the printability of the lattices. At higher relative densities, the material tends to form clumps, trapping uncured resin in the cell cavities, and if relative density is lower than \SI{7}{\percent}, the structure is unable to support itself against gravity. 

  %     \begin{figure}[tb]
  %     \centering
  % \begin{subfigure}[t]{\columnwidth}
  %   \includegraphics[width=\linewidth]{Pics/Inductance_Limit.png}
  %   \caption{}
  %   \label{sfig:comp}
  % \end{subfigure}
  % %--- Second subfigure ---
  % \begin{subfigure}[t]{\columnwidth}
  %   \includegraphics[width=\linewidth]{Pics/EMfield.png}
  %   \caption{}
  %   \label{sfig:typcur}
  % \end{subfigure}
  %     \caption{Fig3}
  %     \label{figure:thickmotive}
  %  \end{figure}

% \begin{figure}[tbp]
%   % \centering
%   % %--- First subfigure ---
%   % \begin{subfigure}[t]{\columnwidth}
%   %   \includegraphics[width=\linewidth,height=0.35\textheight,keepaspectratio]{Pics/Inductance_Limit.png}
%   %   \caption{}
%   %   \label{sfig:comp}
%   % \end{subfigure}

%   % %--- Second subfigure ---
%   % \begin{subfigure}[t]{\columnwidth}
%   %   \includegraphics[width=\linewidth,height=0.35\textheight,keepaspectratio]{Pics/EMfield.png}
%   %   \caption{}
%   %   \label{sfig:typcur}
%   % \end{subfigure}
%       \centering
%         \includegraphics[width=\columnwidth]{Pics/Inductance_Limit_ver2.png}
%   \caption{Fig3}
%   \label{figure:thickmotive}
% \end{figure}

\section{RESULTS AND DISCUSSION}
%%**-----------------------------------------------------------------------
% Characterization method
% Force-inductance characteristics
% Motion demo
% Compare different lattice groupings

%%**----------------------------------------------------------------------
\subsection{Characterization}

The COPESS is characterised on a three-axis indentation setup. Two micrometric manual linear stages control the positions of the X and Y stages. An M-111.1DG translation stage is positioned along the Z-axis above the X stage, which is controlled by the paired C-884.4DC motion controller (Physik Instrumente, USA). An ATI Nano17 (ATI Industrial Automation, USA) load cell is mounted on the Z stage. A plastic tape is placed on the tip of the metallic load cell (used as the indenter) to avoid interference with the ferromagnetic target.  The inductance data is captured via the I2C protocol on the Arduino at a rate of 20 cycles per second using the open-source software CoolTerm. The force data from the load cell is recorded with the LabVIEW GUI. Both the inductance and force data are recorded at 20 Hz. To synchronize the data, each signal is first converted to a time-based data structure using its timestamp, and then synchronized using the nearest-neighbor interpolation method. This effectively aligns signals with different sampling rates or start times. This method provides an approximation that is reasonably precise for comparative analysis and preserves the relative timing accuracy between the two systems.

 % \begin{figure}[tb]
 %      \centering
 %        \includegraphics[width=\columnwidth]{Pics/Setup figure ver2.png}
 %      \caption{The three-axis indentation setup used for sensor characterization. A high-precision motorized translation stage controls the indentation depth, while an ATI Nano17 load cell measures the applied force. The sensor's response is recorded with arduino interface.}
 %      \label{figure:setup}
  
 %   \end{figure}

% Figure … presents the output curve for a lattice structure with a 3 mm cell size, 12 mm height, and 7\% relative density. , while the force required for compression is comparatively lower.  
% Similarly, the same test is performed for lattice structures with 10\% and 20\% relative density, as shown in Figure … and Figure …, respectively. Figure … illustrates that while the change in inductance is smaller, the force range is larger. Furthermore, in Figure …, the inductance variation decreases further, but the force range extends up to 15 N. This demonstrates a trade-off between sensitivity and force range.

Preliminary tests characterizing various cell sizes (\SI{2.2}{\milli \meter}, \SI{3}{\milli \meter} and \SI{4}{\milli \meter}) and relative densities (\SI{7}{\percent}, \SI{10}{\percent}, and \SI{20}{\percent} ) within the available design space (based on printability limits) show that relative density is the biggest factor affecting the stiffness of a given lattice. Among the cell sizes tested, \SI{3}{\milli \meter} showed a superior force range exceeding \SI{2.2}{\milli \meter} by over \SI{70}{\percent} and \SI{4}{\milli \meter} by \SI{35}{\percent} while maintining similar sensitivity. Relative density refers to the percentage of the bounding volume containing the elastic material, with a smaller relative density resulting in a lattice of lower stiffness and vice versa. The gyroidal lattices are varied across three different relative densities and are subjected to cyclic compression tests. Specifically, the lattices of cell size \SI{3}{\milli \meter} and relative densities of \SI{7}{\percent}, \SI{10}{\percent}, and \SI{20}{\percent}  are chosen as they exhibited good repeatability across various prints. The sensorized tile is covered with lattice structures of uniform thickness. The lattice structure is compressed to \SI{50}{\percent} of its original height (\SI{12}{\milli \meter}), and the corresponding change in inductance is recorded. 
% Beyond \SI{50}{\percent} compression, the lattice approaches the densification region, and does not translate to reliable sensor readings.  

\begin{figure}[tb]
  \centering

  \begin{subfigure}[t]{0.5\linewidth}
    \includegraphics[width=\linewidth,height=0.20\textheight,keepaspectratio]{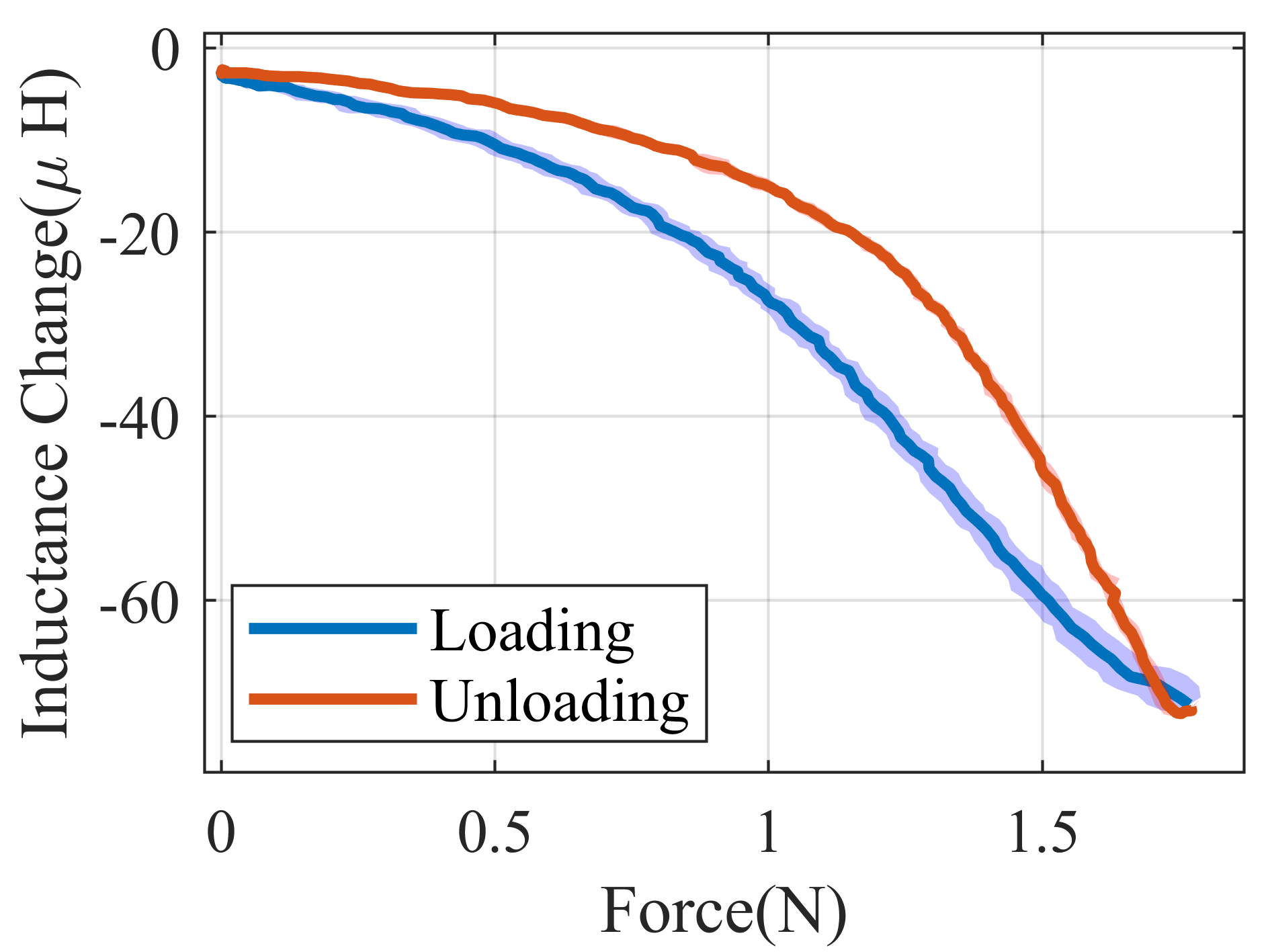}
    \caption{}
    \label{sfig:LvsF7}
  \end{subfigure}

  % \vspace{0.6em}

  \begin{subfigure}[t]{0.5\linewidth}
    \includegraphics[width=\linewidth,height=0.20\textheight,keepaspectratio]{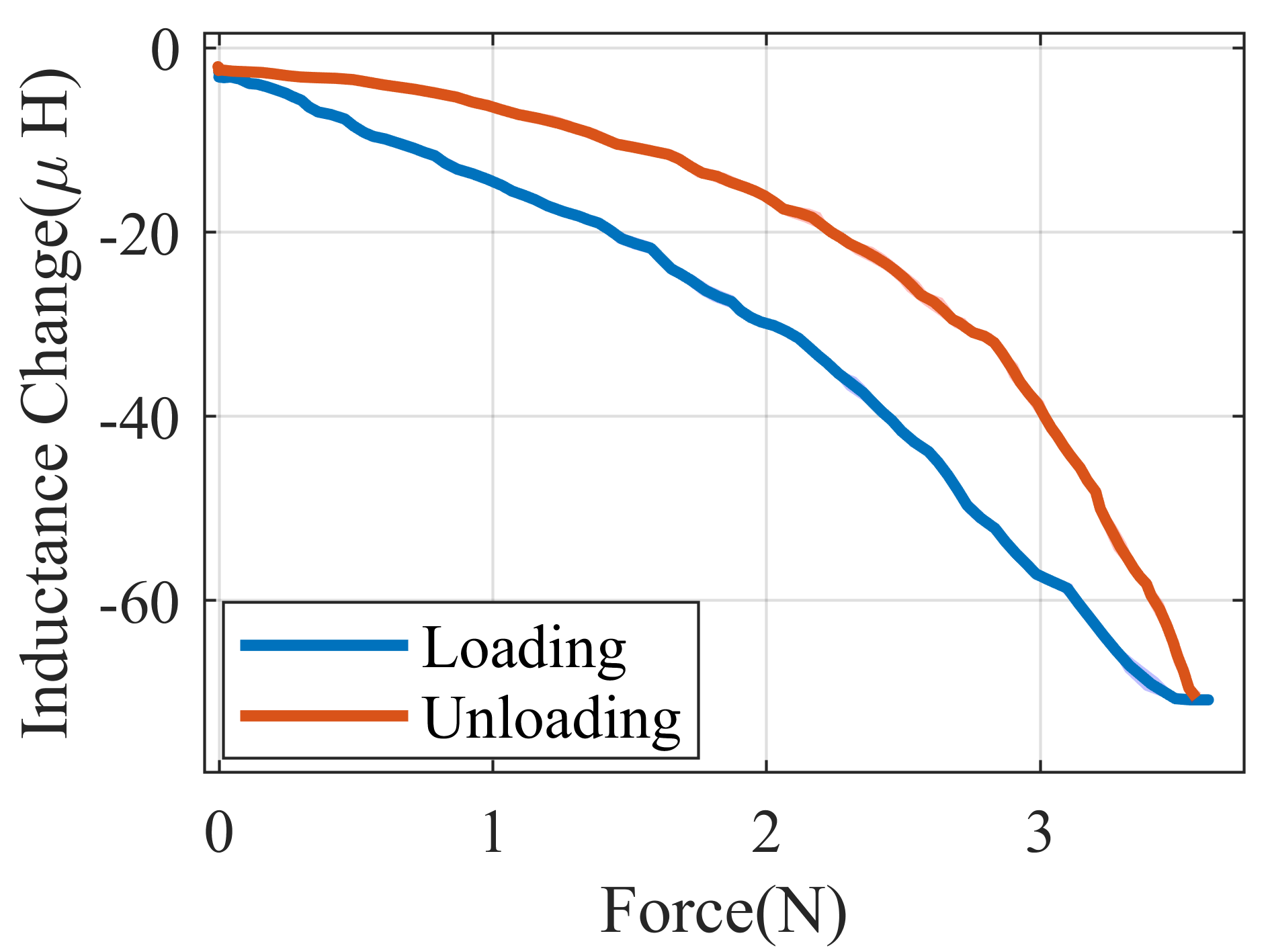}
    \caption{}
    \label{sfig:LvsF10}
  \end{subfigure}

  % \vspace{0.6em}

  \begin{subfigure}[t]{0.5\linewidth}
    \includegraphics[width=\linewidth,height=0.20\textheight,keepaspectratio]{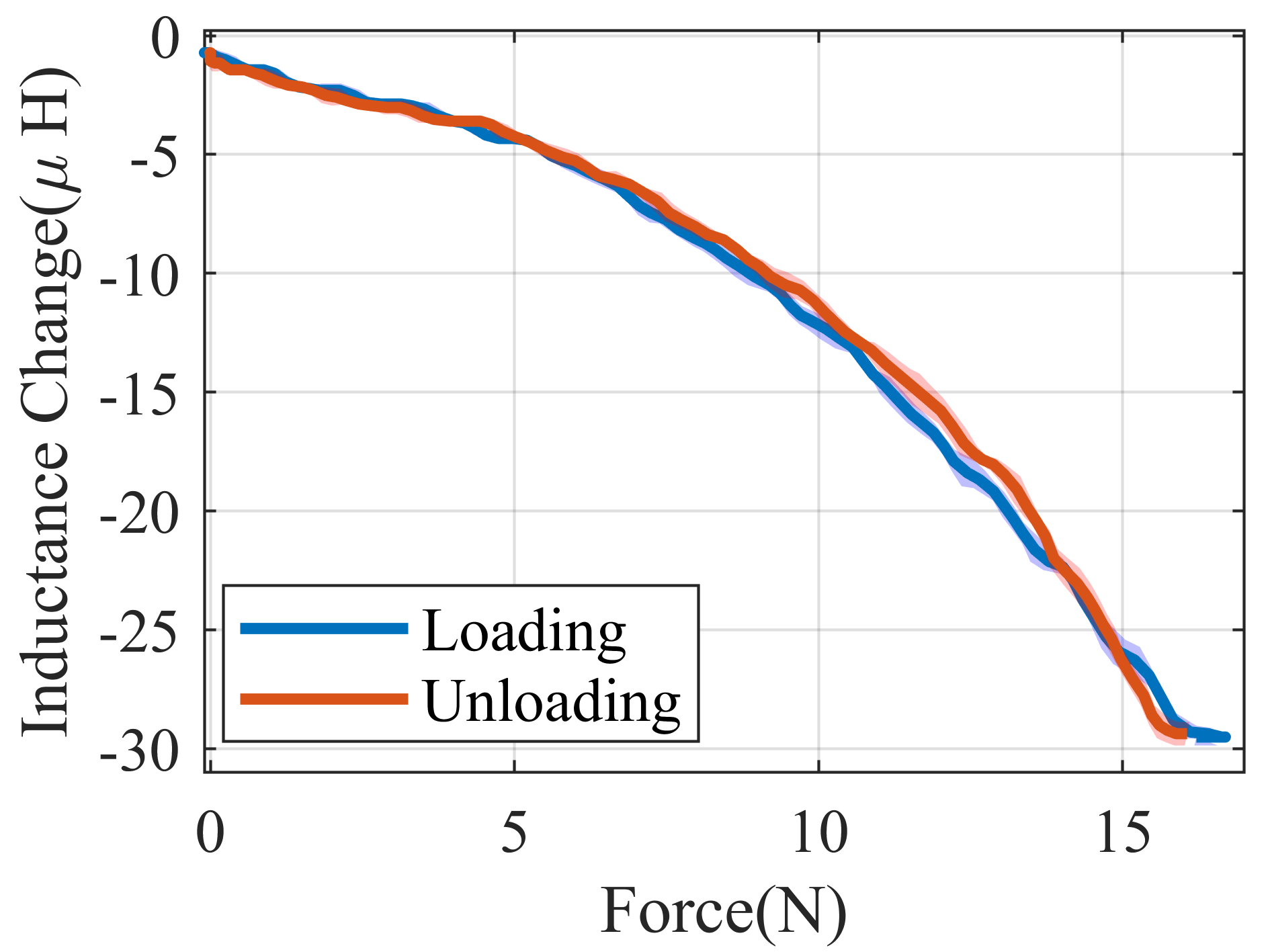}
    \caption{}
    \label{sfig:LvsF20}
  \end{subfigure}

  \caption{Quantitative results demonstrating the co-design of mechanical and sensing properties by tuning lattice relative density. Inductance-force curves for (a) \SI{7}{\percent}, (b) \SI{10}{\percent}, and (c) \SI{20}{\percent} relative densities, illustrating how a higher relative density increases the operational force range at the cost of sensitivity (the slope of the curve). The data are presented as mean and confidence interval.}
  \label{figure:charac}
\end{figure}

\Crefrange{sfig:LvsF7}{sfig:LvsF20}, show the inductance vs force curves of the selected lattices. \Cref{tab:sensordata} shows the resultant sensor characteristics from these configurations. The effective stiffness is calculated from the slope of the linear region (up to \SI{2}{\milli \meter} displacement) of the force-displacement curves. The operational force range is defined as the maximum force measured at \SI{6}{\milli \meter} compression. The sensitivity is calculated as the average slope of the inductance-force curves. The hysteresis is defined as the maximum difference between the loading and unloading curves expressed as a percentage of the full-scale inductance change. Increasing the relative density from \SI{7}{\percent} to \SI{20}{\percent} results in a 7-fold increase in stiffness and a 9-fold increase in operational force range, at the cost of a 23-fold decrease in sensitivity. This demonstrates a predictable design trade-off that allows for tailoring a sensor for specific applications. For applications requiring the delicate detection of lightweight objects like soft fruits, a low relative density (e.g., \SI{7}{\percent}) is ideal due to its high sensitivity. Conversely, for handling heavier, more robust objects, a higher relative density (e.g., \SI{20}{\percent}) provides the necessary force range and structural support. However, the results tend to vary over different lattices of the same characteristics due to minor printing defects.

It can also be observed that the mean values of the loading and unloading curves exhibit low hysteresis (\SI{8.7}{\percent} for \SI{20}{\percent} relative density), which indicates good repeatability and improved accuracy during cyclic loading. 

\begin{figure}[tbph]
  \centering
  %--- Third row (1 small + 1 wide) ---
  \begin{subfigure}[t]{0.35\textwidth}
    \includegraphics[width=\linewidth,height=0.40\textheight,keepaspectratio]{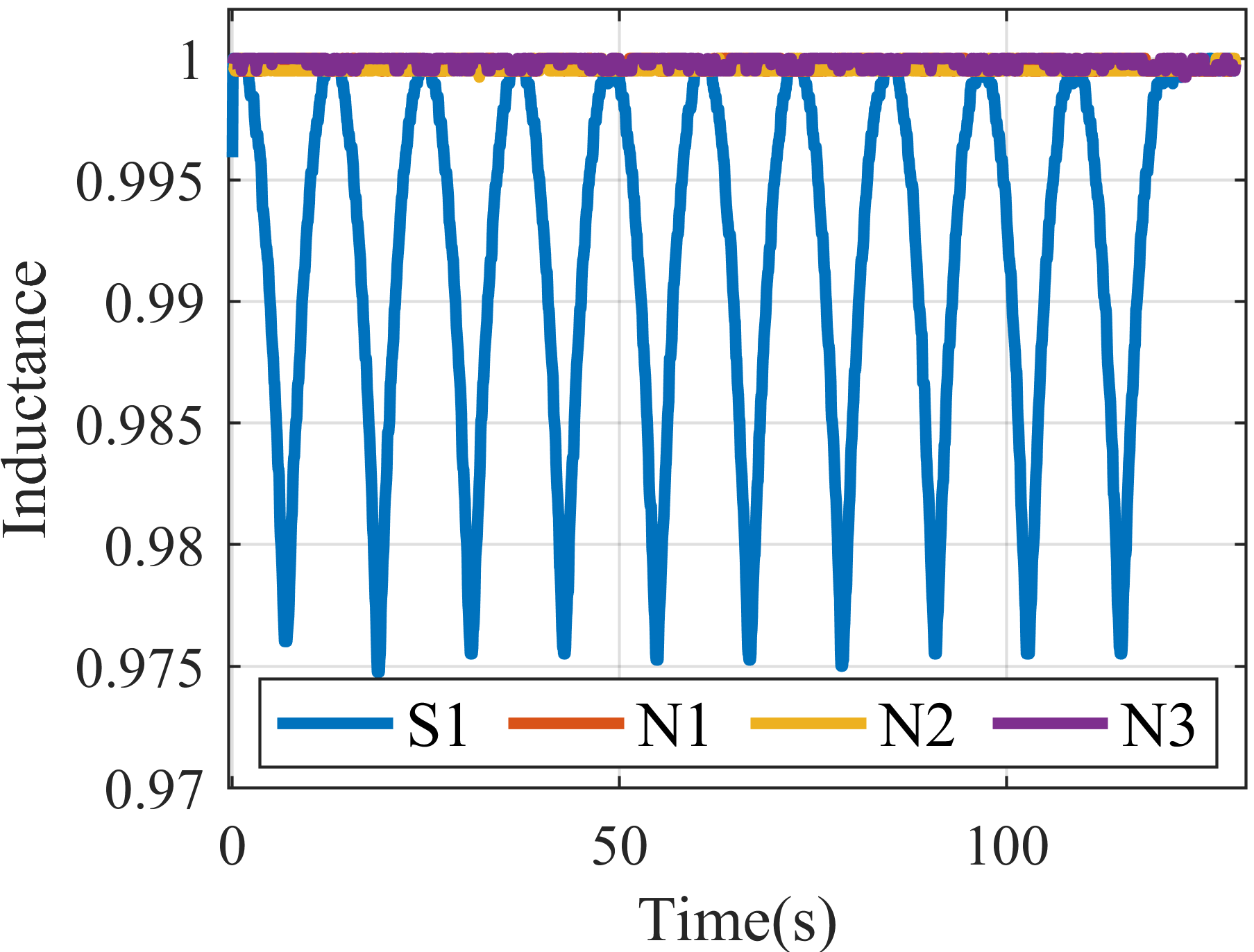}
    \caption{}
    \label{sfig:EMCross}
  \end{subfigure}\hfill
  \begin{subfigure}[t]{0.38\textwidth}
    \includegraphics[width=\linewidth,height=0.40\textheight,keepaspectratio]{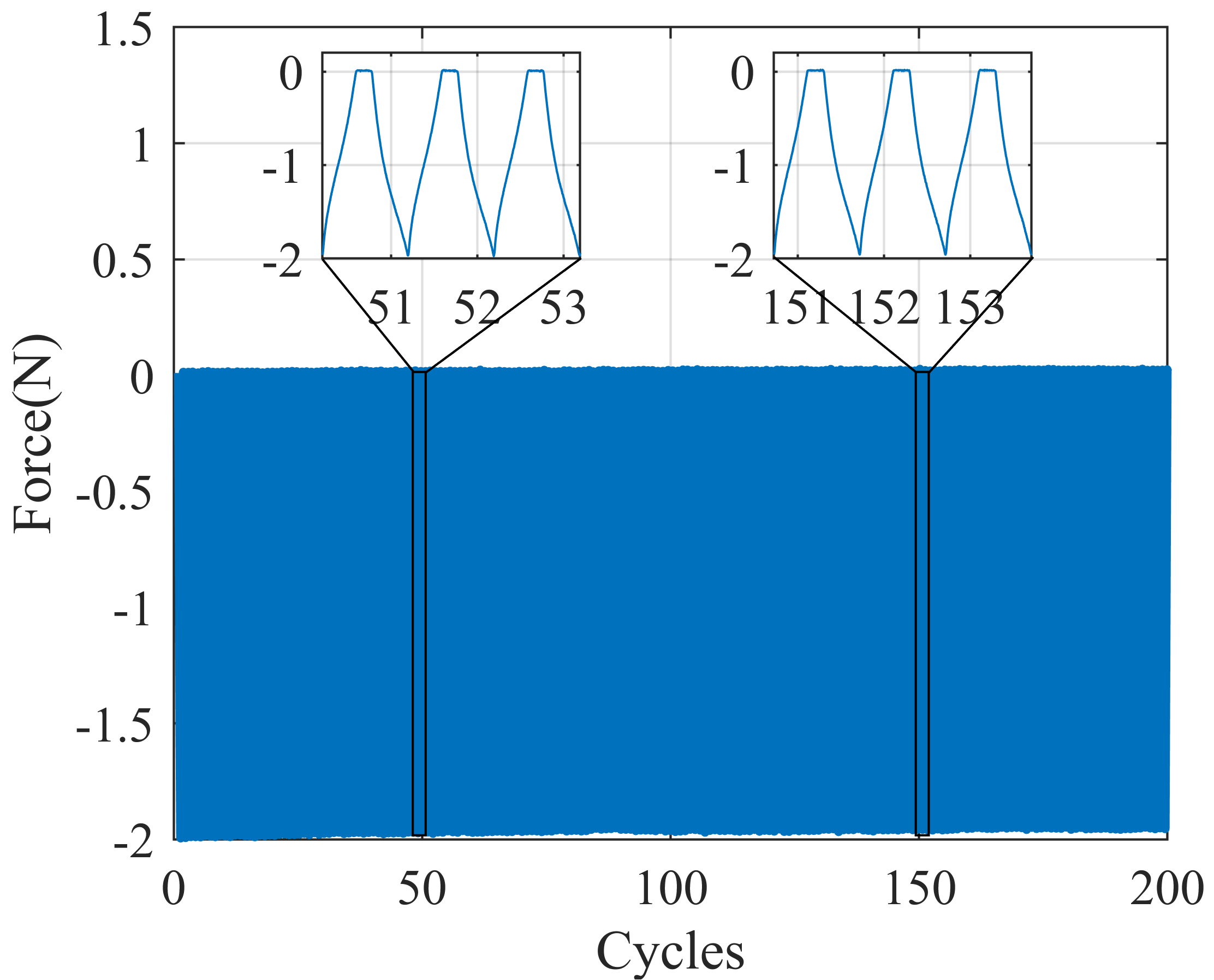}
    \caption{}
    \label{sfig:repeat}
  \end{subfigure}

  \caption{Experimental validation of the sensor's robustness. (a) \ac{EM} crosstalk when a linear vertical force is applied on S1 and the corresponding sensor response from its three neighbouring sensors. (b) Results from a 200-cycle mechanical repeatability test. The \SI{99.96}{\percent} overlap between the first three and last three cycles (inset) demonstrates the high durability and minimal plastic deformation of the 3D-printed gyroid structure.}
  \label{figure:rep}
\end{figure}

When sensors are positioned close to each other, there is a possibility of crosstalk, where the signal from one sensor interferes with its neighbours. To verify this, one sensor is activated while simultaneously recording the data from the adjacent sensors. \Cref{sfig:EMCross} shows no measurable effect on the neighbouring sensors, confirming that sensors don't have an \acs{EM} crosstalk. This demonstrates the electrical robustness of our design. 

% Repeatability of the copess layer is very important for reliable sensor performance and for the reuse of the sensors. 
Lastly, the lattice sensors exhibit high repeatability in both inductance variation and mechanical characteristics. \Cref{sfig:repeat} shows the results from cyclic testing over 200 cycles. The results at the beginning and end of the tests have a \SI{99.96}{\percent} correlation overlap. The high repeatability of the sensor response demonstrates the mechanical robustness and durability of the 3D-printed gyroid structure. The material shows minimal degradation or plastic deformation, validating its suitability for applications involving repeated contact and loading.

  %  \begin{figure*}[tb]
  %     \centering
  % %--- First subfigure ---
  % \begin{subfigure}[t]{0.3\textwidth}
  %   \includegraphics[width=\linewidth]{Pics/FvsdL7(A4).png}
  %   \caption{}
  %   \label{sfig:comp}
  % \end{subfigure}
  % %--- Second subfigure ---
  % \begin{subfigure}[t]{0.3\textwidth}
  %   \includegraphics[width=\linewidth]{Pics/FvsdL10(A6).png}
  %   \caption{}
  %   \label{sfig:typcur}
  % \end{subfigure}  
  %   %--- Third subfigure ---
  % \begin{subfigure}[t]{0.3\textwidth}
  %   \includegraphics[width=\linewidth]{Pics/FvsdL20(A4).png}
  %   \caption{}
  %   \label{sfig:typcur}
  % \end{subfigure} 
  
  %   %--- Fourth subfigure ---
  % \begin{subfigure}[t]{0.3\textwidth}
  %   \includegraphics[width=\linewidth]{Pics/LvsFMean7(A1).png}
  %   \caption{}
  %   \label{sfig:comp}
  % \end{subfigure}
  % %--- Fifth subfigure ---
  % \begin{subfigure}[t]{0.3\textwidth}
  %   \includegraphics[width=\linewidth]{Pics/LvsFMean10(A6).png}
  %   \caption{}
  %   \label{sfig:typcur}
  % \end{subfigure}  
  %   %--- Sixth subfigure ---
  % \begin{subfigure}[t]{0.3\textwidth}
  %   \includegraphics[width=\linewidth]{Pics/LvsFMean20(A5).png}
  %   \caption{}
  %   \label{sfig:typcur}
  % \end{subfigure} 
  
  % %--- Seventh subfigure ---
  % \begin{subfigure}[t]{0.3\textwidth}
  %   \includegraphics[width=\linewidth]{Pics/EMCrosstalk.png}
  %   \caption{}
  %   \label{sfig:typcur}
  % \end{subfigure}  
  %   %--- Eighth subfigure ---
  % \begin{subfigure}[t]{0.6\textwidth}
  %   \includegraphics[width=\linewidth]{Pics/Repeatability.png}
  %   \caption{}
  %   \label{sfig:typcur}
  % \end{subfigure} 
  
  %    \caption{}
  %     \label{figure:charac}
  %  \end{figure*}

\begin{figure*}[h!]
    \centering
        
    %--- First row ---

         \includegraphics[width=0.85\textwidth]{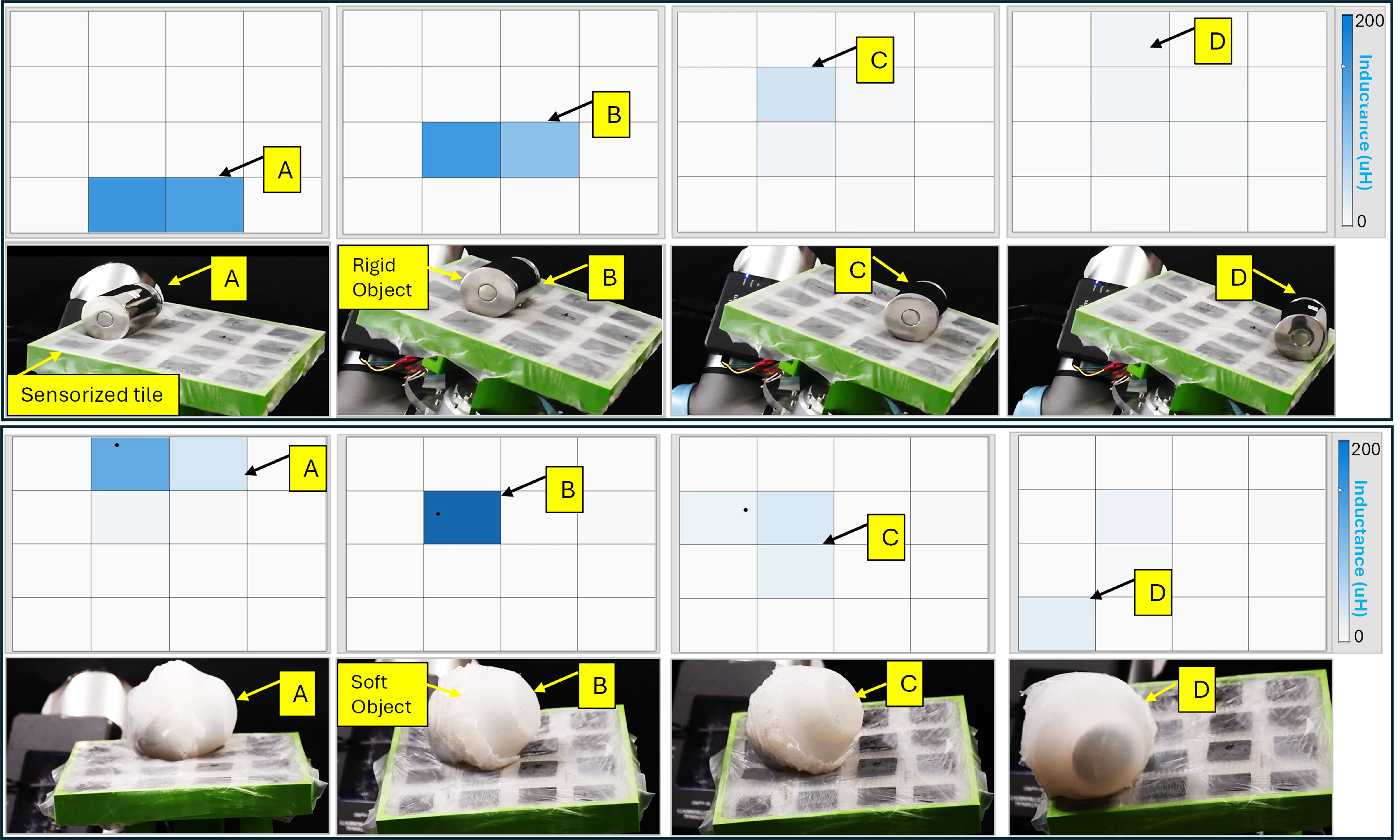}

    \caption{Localized sensing and passive guidance of a manipulated object. The object is moved across the sensorized surface as it transitions from a low-stiffness region to a high-stiffness region, with its position tracked in a LabVIEW GUI. Two types of objects are demonstrated: Top Row: Rigid standard weight, Bottom Row: Irregular soft object}
    \label{fig:demop1}
\end{figure*}

    \begin{table*}
        \centering\
        \resizebox{\textwidth}{!}{%
        \begin{tabular}{|c|c|c|c|c|}
        \hline
            Lattice Relative Density (\%) & Effective Stiffness (N/mm) & Operational Force Range(N) & Sensitivity ($\mu$H/N)  & Hysteresis (\%) \\ \hline
            7 & 0.37 & 1.78 & 39.04 & 20.07\\ \hline
            10 & 0.54 & 3.61 & 18.92 & 17.00\\ \hline
            20 & 2.52  & 16.68 & 1.70 & 8.70 \\ \hline
        \end{tabular}
        }
        \caption{Quantified Performance Metrics Demonstrating the Co-Design of Mechanical and Sensing Properties. Each value represents the mean of N = 5 cyclic compression tests per lattice configuration.}
        \label{tab:sensordata}
    \end{table*}

\subsection{Passive Object Guidance}
% describe the arrangement of lattices on the tile
% Describe the platform we use
% describe the target object
% describe how the test is conducted
% describe how the results are collected
% Describe the results
% Interpret the results
% Discuss the influences mentioned in the first part
% Cite supporting media

\begin{figure}
    \centering
    %--- Second row ---
    \begin{subfigure}[t]{0.35\textwidth}
        \includegraphics[width=\linewidth,height=0.40\textheight,keepaspectratio]{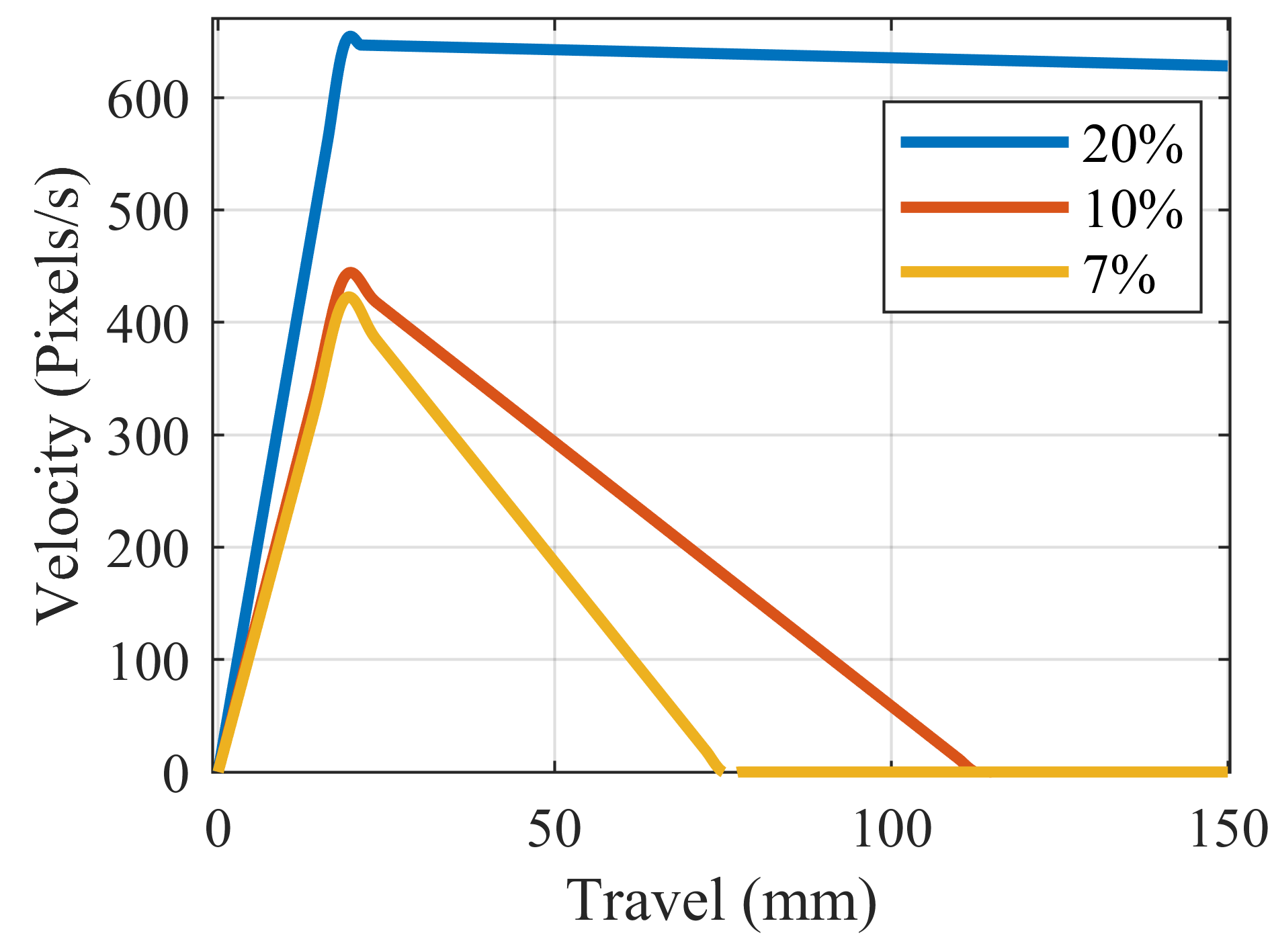}
        \caption{}
        \label{sfig:VelInd}
    \end{subfigure}
    \hfill
    \begin{subfigure}[t]{0.35\textwidth}
        \includegraphics[width=\linewidth,height=0.40\textheight,keepaspectratio]{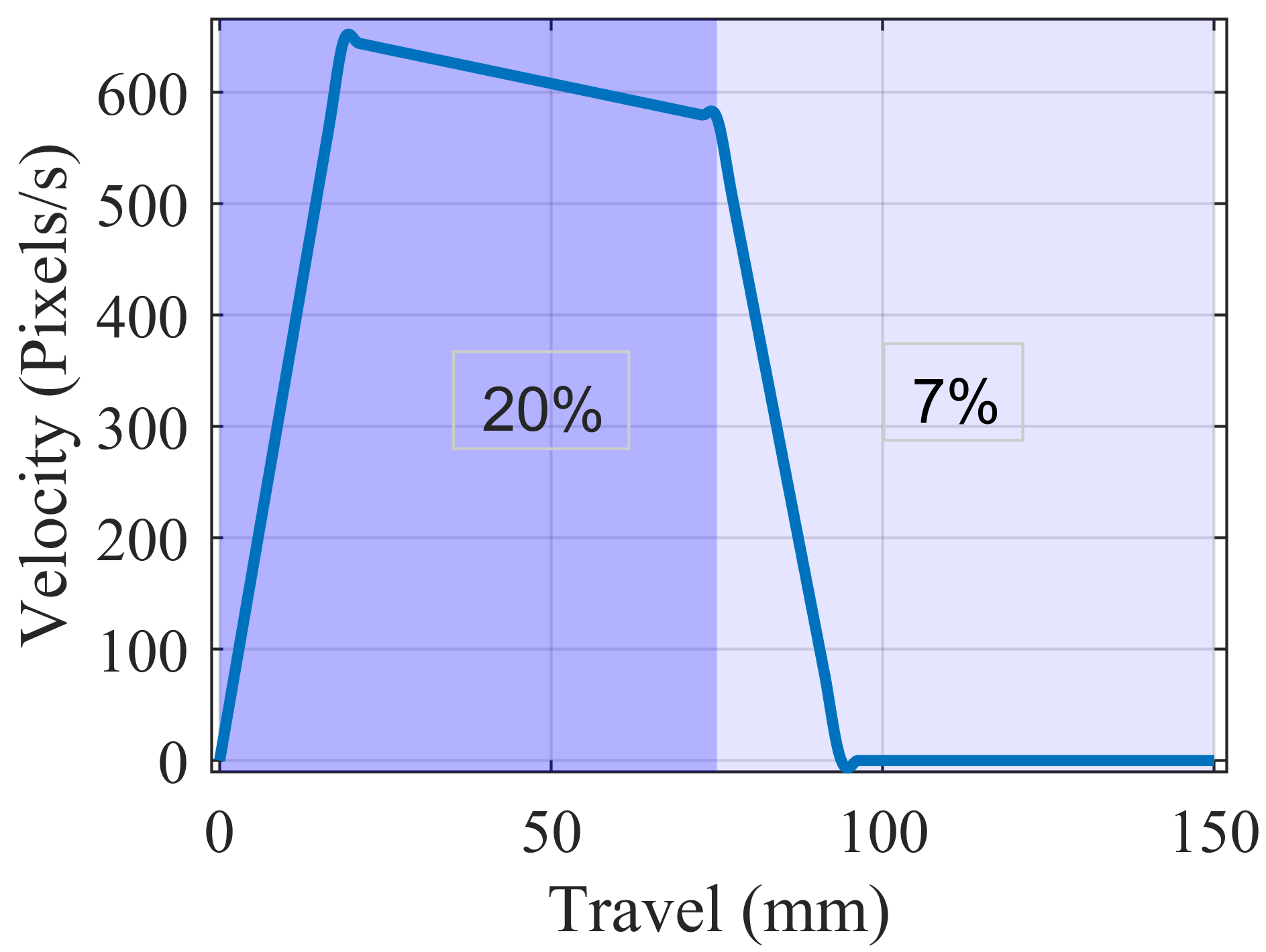}
        \caption{}
        \label{sfig:transvel}
    \end{subfigure}
\caption{ Object velocity and travel distance on tiles with varying relative densities. (a) The \SI{20}{\percent} region, being stiffer, exhibits lower friction, allowing continuous motion, whereas lower relative densities increase friction and cause faster deceleration. (b) Passive guidance of the object across regions with different relative densities. In the \SI{20}{\percent} region, the object maintains motion with minor velocity drops, while in the \SI{7}{\percent} region, it comes to rest rapidly due to higher friction.}
   \label{fig:demop2}
\end{figure}

Finally, i) the influence of these topologies on sensing in a moving platform; ii) the combination of different lattices on a sensor tile and its effect on sensing and motion of a given target object, are investigated. The tile is fixed to the end of a robotic arm (UR5e, Universal Robots, Denmark) using a custom 3D printed frame. The UR5 tilts the tile in a controlled manner. To showcase \acs{COPESS}’s sensing ability in versatile conditions, two different objects are manipulated: A rigid standard \SI{500}{\gram} weight and an irregularly shaped soft object cast with silicon.    
In the first set of experiments, the tile is covered with lattice structures of relative density (\SI{7}{\percent}) and tilted to \SI{\pm 20}{\degree} at \SI{0.2}{\degree / \second} so that the object can move from one edge of the tile to the other. The sensor responses are recorded in a LabVIEW GUI, as shown in the video at 0:54. At lower inclination angles, the object remains stable on the sensor and highly compresses the lattice, resulting in a more pronounced sensor response. As the angle increases, however, the object begins to roll, gaining speed. When passing over the sensors at higher velocities, the object has less time to compress the lattice, leading to a weaker response. Nevertheless, the sensor resolution is sufficient to determine the object's position. This experiment thus validates the soft surface's capability to provide localised sensing of objects during manipulation (\Cref{fig:demop1}) and, eventually, to retrieve the object's velocity from the sensors' responses. Additionally, the results show that the sensing response is only dependent on the objects' weight and not affected by their shape or stiffness (which, on the other hand, affect the movement trajectory).\\

In the next set of experiments, the tile is covered with lattices with a \SI{7}{\percent} relative density on half of its area and the other half with a \SI{10}{\percent} relative density. The object is then moved from one edge of the tile to the other, as in the previous tests. When the object enters the \SI{7}{\percent} region, it moves more slowly because the lower stiffness leads to larger compressions due to the object’s load. Greater compression also results in a stronger sensor response. In contrast, when the object transitions to the \SI{10}{\percent} region, the object moves more quickly, and the sensor response is weaker than in other sections. 
To further demonstrate passive object guidance, the object is given an initial impulse manually, and an overhead camera records the object’s motion as shown in the video at 1:19. The surface is covered with each relative density combination individually, and the velocity and the distance traveled are obtained using an OpenCV script. \Cref{sfig:VelInd} shows the velocity plots along three different lattice stiffness regions.   
 As the object rolls over the \SI{20}{\percent} region, the velocity is maintained with a slow deceleration. When the object interacts with the \SI{10}{\percent} region, it gains acceleration initially but eventually starts decelerating and stops after traveling approximately \SI{110}{\milli \meter}. On the other hand, in the \SI{7}{\percent} region, the object decelerates after covering a short distance (\SI{75}{\milli \meter}) because of higher compliance. 
\Cref{sfig:transvel} shows the same experiment on a tile covered with lattices of different stiffness.
On transitioning from a \SI{20}{\percent} region to a \SI{7}{\percent} region, the velocity drops suddenly. 
%\Cref{sfig:transvel} shows that when the object rolls in the \SI{20}{\percent} region, the velocity drop is very small because of low friction (object covers ~\SI{80}{\milli \meter}), but as it enters the \SI{7}{\percent} region, the velocity drops suddenly, and the object stops moving due to high friction (covering ~\SI{15}{\milli \meter}). This experiment demonstrates that the surface stiffness gradient can be used to passively control and guide object motion.%

The influence of the lattice stiffness on the required actuation effort is also evident. To initiate motion on the stiffest uniform surface (20\% relative density), a tilt angle of only 5° is required, whereas the much softer 7\% surface needed a significantly larger tilt angle of 12.7° to overcome the higher initial stiction and rolling resistance, quantitatively demonstrating the direct link between programmed stiffness and the energy required for manipulation. These findings confirm that the motion of an object can be modulated by varying the stiffness of the \acs{COPESS} layer, thereby highlighting the potential to passively guide objects in specific directions for practical applications, such as food packaging. Overall, the results offer promising avenues to design motion pathways that control an object's trajectory and relative velocity while improving actuator energy efficiency.
   
     %  \begin{table}[h!]
%\centering
% \resizebox{\columnwidth}{!}{%
% \begin{tabular}{|c|c|c|c|}
% \hline
%  Case & Infill \% Stiffness Gradient& Tilt Angle (°) & Rolling Speed (cm/s)  \\ \hline
% 1&  7\% (soft) – full tile& 12.69 ↑ (largest) & ↓ (slowest) \\ \hline
% 2&  10\% → 7\% (softening)&  8.95 Medium–High & Medium–Low \\ \hline
%  3& 20\%-7\% (softening)&  7.33&   High \\ \hline
%  4& 20\% Uniform stiff & 5 ↓ (slowest) & ↑ (Highest) \\ \hline
% \end{tabular}%
% }
% \caption{Different lattice configurations with the tilt angle and speed of the object}
% \label{tab:velocitydata}
% \end{table}
 
\section{CONCLUSIONS}
In this study, a \ac{COPESS} integrated with an inductive sensor array is proposed for use in soft surface manipulation. Gyroidal lattices printed via \acs{SLA} form the deformable layer of the inductive sensor. The lattices are tuned by changing their relative density. The experiments demonstrate that by tuning the lattice stiffness, sensors with varying force ranges and sensitivities can be designed. The presented sensing platform demonstrates consistent and repeatable performance, even during the manipulation of soft and irregular objects, while maintaining a strong sensing response over a wide force range. Moreover, we employ the inductive sensor as a force sensor, which lead to a larger inductance variation over a broad force range compared to previously reported inductive sensors, either using silicones \cite{wang2019wireless} or 3D printed lattices \cite{bhaumik2025inductive} as modulating layers. The design of lattice structures is constrained by printability limitations and sensor performance. More efforts will be needed to extend the force working range while preserving high sensitivity.  A more extensive study of other lattice structures and materials can be a future avenue of research. Additionally, the system is shown to effectively track an object's position and relative velocity on a moving platform regardless of the objects' characteristics. Lastly, the effect of combining lattices of different stiffnesses on actuation is studied. 
The results presented here offer many promising avenues for the use of lattices in soft handling systems. Future work includes devising a feedback mechanism to aid actuation and actively control the stiffness of the lattice structures. 

\begin{acronym}
\acro{COPESS}{COmpliant Porous-Elastic Soft Sensing}
\acro{EM}{Electromagnetic}
\acro{PDMS}{Polydimethylsiloxane}
\acro{AC}{Alternating Current}
\acro{TPMS}{triply Periodic Minimal Surfaces}
\acro{SLA}{stereolithography}
\acro{FSR}{force-sensing resistor}
\end{acronym}

\bibliographystyle{IEEEtran}
\bibliography{References}

\end{document}